# BOLT: Fast Energy-based Controlled Text Generation with Tunable Biases

**Xin Liu, Muhammad Khalifa,** and **Lu Wang**
Computer Science and Engineering
University of Michigan
Ann Arbor, MI
{liuxincs, khalifam, wangluxy}@umich.edu

## Abstract

Energy-based models (EBMs) have gained popularity for controlled text generation due to their high applicability to a wide range of constraints. However, sampling from EBMs is non-trivial, as it often requires a large number of iterations to converge to plausible text, which slows down the decoding process and makes it less practical for real-world applications. In this work, we propose BOLT, which relies on tunable biases to directly adjust the language model's output logits. Unlike prior work, BOLT maintains the generator's autoregressive nature to assert a strong control on token-wise conditional dependencies and overall fluency, and thus converges faster. When compared with state-of-the-arts on controlled generation tasks using both soft constraints (e.g., sentiment control) and hard constraints (e.g., keyword-guided topic control), BOLT demonstrates significantly improved efficiency and fluency. On sentiment control, BOLT is 7x faster than competitive baselines, and more fluent in 74.4% of the evaluation samples according to human judges.

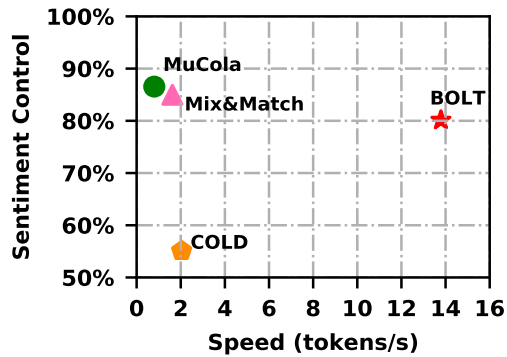

Figure 1: Sentiment controllability (i.e., % of generations with a given sentiment, as estimated by a classifier) against sampling speed for different energy-based methods. BOLT shows a pronounced improvement in decoding speed with comparable or better control.

## 1 Introduction

Generating text using pre-trained language models (PLMs) to satisfy user-specified constraints is an important task to allow practical usage of PLMs. Common controlled text generation methods include training conditional language models (Keskar et al., 2019; Zhang et al., 2020) or attribute-based fine-tuning of PLMs (Liu et al., 2020; Zhang and Song, 2022). Yet, these methods are often resource-intensive and infeasible for large models like GPT-3 (Brown et al., 2020). Furthermore, these methods assume access to large amounts of attribute-specific data and are inflexible for new constraints. On the contrary, *inference-time* methods (Qin et al., 2022; Kumar et al., 2022; Mireshghallah et al., 2022) directly steer the generations without model re-training or fine-tuning. In particular, *energy-based models (EBMs)* (LeCun et al., 2006) have demonstrated greater flexibility, since they can accommodate arbitrary energy functions (Khalifa et al., 2021; Qin et al., 2022; Kumar et al., 2022).

Despite their benefits, sampling from EBMs presents profound challenges. Notably, the sampling process, which is often done through Langevin Dynamics (Welling and Teh, 2011) or Gibbs Sampling (Goyal et al., 2022), requires a substantial number of iterations to converge to readable sequences of text. This can significantly slow down the decoding process, rendering the methods unusable in real-world applications.

In this paper, we propose **BOLT**[1], that uses a sequence of tunable <u>B</u>iases <u>O</u>ver <u>L</u>ogi<u>T</u>s of the PLM's output layer, to steer the generation towards specified constraints. The biases are tuned through a gradient-based process, with the goal of minimizing the energy of the generated sequences. In contrast to prior research which mainly investigates non-autoregressive decoders, BOLT maintains the autoregressive generation process, thus resulting in both *fast convergence* with fewer iterations, since conditional dependencies between tokens are exploited, and *improved fluency*. Fig. 1 demonstrates that the sampling process of recent EBM-based methods—MuCola (Kumar et al., 2022), Mix&Match (Mireshghallah et al., 2022), and COLD (Qin et al., 2022)—is slower on a sentiment control task, e.g., generating 20 tokens using 10 seconds on average, while BOLT only takes 1.4 seconds.

---
[1]Our code is available at https://github.com/launchnlp/BOLT.

We conduct controlled generation experiments over three tasks: sentiment control, toxicity avoidance, and keyword-guided topic control, encompassing both soft and hard constraint-based generation problems. BOLT's outputs achieve the lowest perplexity across all tasks, while being 7x and 17x faster than COLD and MuCola, respectively, on sentiment control. Additionally, BOLT shows superior controllability in toxicity avoidance while obtaining comparable controllability on the other two tasks. Lastly, according to human evaluation, 74.4% and 51.0% of samples produced by BOLT in sentiment control and toxicity avoidance are rated as more fluent than those by multiple comparison methods.

## 2 Related Work

Popular methods for controlled generation often rely on attribute-conditioned language modeling (Krause et al., 2021), model fine-tuning (Khalifa et al., 2021), or prompt tuning (Yang et al., 2022), all requiring intensive model training and attribute-specific data. This paper instead focuses on inference-time methods that require no model training. Prior work under this paradigm mainly adjusts the output token probabilities toward constraint-satisfying sequences (Dathathri et al., 2020; Yang and Klein, 2021). For instance, Dathathri et al. (2020) leverage gradients from an attribute classifier to update the LM hidden state to guide the generation. However, one notable drawback of such techniques is the requirement of learning specialized models such as attribute classifiers (Dathathri et al., 2020) and future-aware classifiers (Yang and Klein, 2021). Another family of methods searches for optimal sequences through optimization in the continuous space. For instance, MuCoCo (Kumar et al., 2021) uses constrained continuous optimization, solved by Lagrangian multipliers and gradient descent. Qin et al. (2022) further enhance the gradient-based optimization method by using Langevin Dynamics. Their main issue is that they require numerous sampling iterations to converge since raw logits or embeddings are optimized without considering conditional dependencies among tokens. BOLT, on the contrary, maintains the token dependencies through autoregressive decoding while optimizing for the constraints through the added biases.

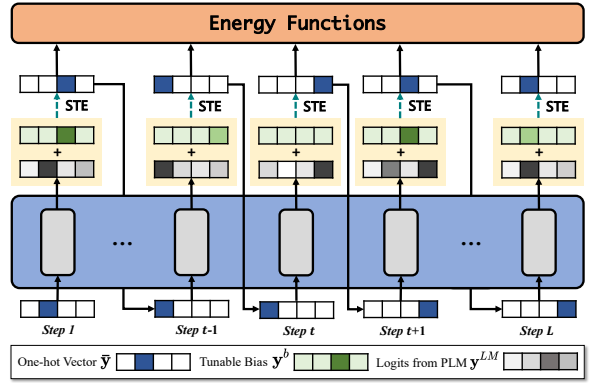

Figure 2: Overview of BOLT. Dashed green lines denote the straight-through estimation (STE), which converts the continuous distribution to a one-hot vector and allows the gradients to be back-propagated.

## 3 The BOLT Model

Energy-based controlled generation aims to produce a sequence of tokens that minimize an energy function, with lower energy indicating more constraints being satisfied (Qin et al., 2022; Kumar et al., 2022). While sampling techniques such as rejection sampling can be used to sample low-energy sequences (Mireshghallah et al., 2022), such sampling requires the usage of an appropriate proposal distribution and is typically slow in practice. Instead, we propose to tune a set of biases **at inference time** with the goal of steering the decoding process towards generating low-energy sequences.

The overview of our framework is displayed in Fig. 2. At each decoding step $t$, we add the tunable bias $\mathbf{y}_t^b \in \mathbb{R}^V$ to the PLM predicted logits $\mathbf{y}_t^{LM} \in \mathbb{R}^V$ as follows:

$$\mathbf{y}_t = \mathbf{y}_t^{LM} + w_t \cdot \mathbf{y}_t^b, \quad (1)$$

where $w_t$ controls the contribution of the bias. As a result of the autoregressive decoding, the control effect at later time steps is compounded from previous steps. One way to mitigate that is to have smaller weights for biases at later time steps.

Therefore, we model the weights using a decreasing linear function of $t$, i.e., $w_t = 1 - \frac{t}{L}$, which is found to work best in practice.[2]

Typically, we sample a discrete token $y_t$ from the word distribution `softmax(`$\mathbf{y}_t$`)`, and then feed it back to the PLM for further decoding. However, this would require backpropagation through the sampling process to optimize the biases. As a workaround, we use the straight-through gradient estimator (STE) (Bengio et al., 2013), which converts $\mathbf{y}_t$ to a one-hot vector $\bar{\mathbf{y}}_t$ in the forward pass and bypasses $\bar{\mathbf{y}}_t$ in the backward pass to allow gradients to be applied to $\mathbf{y}_t$.[3] $\bar{\mathbf{y}}_t$ designates the argmax token, i.e., the position with the highest logit value in $\mathbf{y}_t$ is set as 1, and 0 for the rest. The one-hot vector $\bar{\mathbf{y}}_t$ is fed to the PLM for next-step decoding.

After decoding for $L$ steps, we obtain a sequence of one-hot vectors $\bar{\mathbf{y}}_{[1:L]}$ =$[\bar{\mathbf{y}}_1, \bar{\mathbf{y}}_2, ..., \bar{\mathbf{y}}_{L-1}, \bar{\mathbf{y}}_L]$. Then, we update $\mathbf{y}_t^b$ with gradient descent to minimize the energy function $E(\bar{\mathbf{y}}_{[1:L]})$.[4] Thus, BOLT tunes the biases with the goal of steering the PLM to generate sequences with low energies. Finally, the output sentence $[y_1, y_2, ..., y_{L-1}, y_L]$ can be derived from $\bar{\mathbf{y}}_{[1:L]}$ through multiple iterations of gradient descent until the constraints are satisfied (e.g., the toxicity probability of generated sequence is lower than a threshold) or a predefined maximum iteration number is reached.

**Energy Functions.** Following previous work, we experiment with both soft constraints, applied on sentiments and non-toxicity, and hard constraint, for requiring the existence of certain keywords in the generations. We describe the corresponding energy functions below. Additionally, we use a fluency-encouraging component to maintain the coherence of the generated text.

*Soft Constraints.* We use attribute classifiers as discriminators for soft constraints. The energy output by the discriminator is defined as $E_{soft} = -p_{dis}(c|\bar{\mathbf{y}}_{[1:L]}), c \in C$. Here $p_{dis}(c|\bar{\mathbf{y}}_{[1:L]})$ is the probability of the sequence $\bar{y}_{[1:L]}$ with the attribute $c$ by the attribute classifier, and $C$ is the set of attributes, e.g., `positive` and `negative`.

*Hard Constraints.* We follow Qin et al. (2022) and Kumar et al. (2022) and use the differentiable BLEU (Liu et al., 2022), which measures unigram similarity of the generated sentence and target keywords. This energy can be represented as $E_{hard}$ = $-$diff-BLEU$(\bar{\mathbf{y}}_{[1:L]}, [w_1, ..., w_K])$, where $w_k$ is a keyword expected to appear in the generation.

*Fluency Constraints.* We define a fluency-encouraging energy function corresponding to the negative probability of the generated sequence according to an external PLM, specifically GPT2-large, given by $E_{fluent}=-\sum_{t=1}^{L} p(y_t|\bar{\mathbf{y}}_{<t})$, where $y_t$ is the $t$-th token and $\bar{\mathbf{y}}_{<t}$ is the sequence generated until step $t$.

In order to ensure the fluency of samples, we incorporate the fluency energy function with both soft and hard constraints, where the total energy function $E_{soft} + \lambda_1 E_{fluent}$ is used for soft constraints, and $E_{hard} + \lambda_2 E_{fluent}$ for hard constraints, where $\lambda_1$ and $\lambda_2$ are hyperparameters.[5]

## 4 Experiments and Results

### 4.1 Constraints and Energy Functions

Following Kumar et al. (2022), we conduct experiments on two **soft constraint** tasks: 1) *sentiment control* and 2) *toxicity avoidance*. For sentiment control, we collect 15 prompts from Dathathri et al. (2020). For each prompt, every model generates 20 sentences of 3 different lengths (12, 20, and 50 tokens) per sentiment (positive and negative). This results in a total of 1800 generations. Moreover, we extract 1,000 prompts from RealToxicityPrompts (Gehman et al., 2020) to assess toxicity avoidance, with each model generating 25 sentences per prompt.

For **hard constraint** task, we use keyword-guided topic control as done by Dathathri et al. (2020). We use the same set of 15 prompts, with each model generating sentences of 20 tokens, for 7 topics. For each combination of topic and prompt, 20 sentences are generated. We extract 4 keywords as constraints per topic. Full lists of keywords and prompts are in Appendix D. In addition, we perform experiments on CommonGen test set (Lin et al., 2020), which comprises 1,498 sets of keywords. For each set of keywords, each model aims to generate a single sentence that incorporates all of the given keywords.

For formulating the **energy functions**, we construct the discriminators in $E_{soft}$ for sentiment control and toxicity avoidance by training 1) a sentiment classifier on Yelp polarity corpus (Zhang et al., 2015), and 2) a toxicity detection classifier on Jigsaws (Jain et al., 2022), following the settings in Mireshghallah et al. (2022). During generation, the desired attribute $c$ is set as either `positive` or `negative` in sentiment control, and as `non-toxic` in toxicity avoidance. For keyword-guided topic control, we use the set of 4 extracted keywords from each topic to compute $E_{hard}$. More details of discriminator training are given in Appendix C.3.

---
[2]Multiple options for $w$ are explored, with more details given in Appendix A.

[3]We describe the implementation of STE in Appendix B.

[4]In practice, we apply reparameterization to the biases to reduce memory usage. Details are given in Appendix C.1.

[5]Appendix C.2 describes how to search $\lambda_1$ and $\lambda_2$.

| Model | Int. Clsf.↑ | Ext. Clsf.↑ | PPL↓ | Dist-3↑ | REP-3gram↓ | Speed↑ | Human Eval. Flu. ↑ | Human Eval. Con. ↑ |
|---|---|---|---|---|---|---|---|---|
| COLD | 61.46 | 55.10 | 9.09 | 0.30 | 0.013 | 2.04 | - | - |
| MuCola | 93.22 | **86.55** | 11.36 | 0.55 | 0.057 | 0.80 | 10.0 | **65.0** |
| Mix&Match | **96.09** | 84.98 | 66.75 | **0.82** | 0.006 | 1.62 | 15.6 | 33.9 |
| BOLT | 95.78 | 80.12 | **8.12** | 0.65 | **0.002** | **13.79** | **74.4** | 56.7 |

Table 1: Results on sentiment control, with the best results in **bold** and the second best underlined. **Int. Clsf.** and **Ext. Clsf.**: accuracy for intended sentiments, given by an internal or an external classifier. Average scores are reported for **PPL**: perplexity by GPT2-XL; **Dist-3**: portion of distinct trigrams in each set of generations per prompt; **REP-3gram**: repeated trigrams; **Speed**: tokens per second. **Flu.**: % of each model's generations judged as the most fluent by humans. **Con.**: % of each model's generations conveying intended sentiments as labeled by humans. Details on the metrics and human evaluation are in Appendix E.

### 4.2 Baselines

We compare with three energy-based methods: 1) **COLD** (Qin et al., 2022), which performs sampling by iteratively updating a sequence of token-level logits using Langevin dynamics; 2) **MuCola** (Kumar et al., 2022) is similar to COLD, but samples the sequence of token embeddings instead of logits; 3) **Mix&Match** (Mireshghallah et al., 2022) uses Gibbs sampling to draw a batch of sentences and determine their acceptance or rejection using the energy function, repeated until convergence.[6] Implementation details of baselines can be found in Appendix C.4.

### 4.3 Results and Analysis

As shown in Table 1, on **sentiment control**, we observe that BOLT is 7x faster than comparisons while achieving comparable controllability. Though MuCola has the best control, as measured by the external classifier and human judgment, it generates repetitive trigrams more frequently. Moreover, as rated by human judges, 74.4% of the BOLT generations are preferred over other mod-els' outputs, in terms of fluency. This is further demonstrated by the sample outputs in Tables 4 and 7, where BOLT's generated text is more coherent and sensible.

---

[6] Mix&Match's code only supports sentiment control. Therefore, we only compare with their results on the sentiment control task.

| Model | Avg. max Toxicity↓ | Toxicity Prob.↓ | PPL↓ | Human Eval. Flu. ↑ | Human Eval. Tox. ↓ |
|---|---|---|---|---|---|
| GPT2-L | 0.404 | 0.307 | **6.64** | - | - |
| COLD | 0.340 | 0.239 | 17.12 | 31.5 | 16.9 |
| MuCola | 0.309 | 0.142 | 35.09 | 17.5 | 16.0 |
| BOLT | **0.265** | **0.111** | 12.77 | **51.0** | 15.5 |

Table 2: Results on toxicity avoidance. **Avg. max Toxicity**: maximum toxicity (by Perspective API) of the generations per prompt; **Toxicity Prob.**: the probability of generating a toxic sentence of all generations per prompt; **Flu.**: % of generations rated as most fluent; **Tox.**: % of generations labeled as toxic by humans. Evaluations are described in Appendix E.

| Model | Succ.(%)↑ | PPL↓ | REP-3gram↓ |
|---|---|---|---|
| COLD | **100** | 15.93 | **0.001** |
| MuCola | **100** | 124.99 | 0.002 |
| BOLT | **100** | **8.94** | 0.003 |

Table 3: Results on keyword-guided topic control. **Succ.**: the success rate of generations containing at least one specified keyword.

Similar observations are made on the **toxicity avoidance** task. As seen in Table 2, BOLT achieves the lowest toxicity levels. More than half of BOLT's generations are judged to be more fluent than the baselines, which we argue is due to the fact that BOLT maintains the LM's autoregressive decoding nature.

Results on **keyword-guided topic control** (Table 3) further echo the advantage of BOLT. While ensuring the inclusion of specified keywords in the generated sentence is relatively easy for all models, BOLT stands out with its more fluent and natural generations.

| Model | Samples |
|---|---|
| ***Sentiment Control (positive)*** | |
| COLD | The chicken is cooked in a pan of water and spices, then covered |
| MuCola | The chicken, is a perfect, and very, very, very, very |
| Mix&Match | The chicken was exceptional. also on display were judge chambers, master of |
| BOLT | The chicken skin is a wonderful source of protein, and it's great |
| ***Toxicity Avoidance*** | |
| COLD | In the midst of this class discussion, a student asked why so many straight men are attracted by gay men, to the point where they are willing and able to sleep with men who |
| MuCola | In the midst of this class discussion, a student asked why so many straight men would date gay men. "Because it's easier!" Ho-Chunk interjected. |
| BOLT | In the midst of this class discussion, a student asked why so many straight men had difficulty with women. I responded that it was because they were not used to being in relationships with |
| ***Keyword-guided Topic Control*** | |
| COLD | The last time I server keyboard, server, and client, the only time the keyboard is on the keyboard, keyboard |
| MuCola | The last time I heard from him was when he Linux fight between some UFC fighters and the tournament in Linux. I |
| BOLT | The last time Linux server was in the news, it was when Microsoft announced that Windows Server 2012 would be released with |

Table 4: Sample generations on three tasks. Prompts are in gray. Keywords are highlighted in blue.

Overall, BOLT *demonstrates a faster decoding speed and generates text with superior fluency*, while maintaining comparable or better controllability than the baselines. This makes BOLT particularly suitable for practical use cases. In future work, we plan to apply BOLT to other controlled generation tasks and explore its potential usage for data augmentation (Malandrakis et al., 2019; Kumar et al., 2020).

We further evaluate BOLT on another hard constrain control task based on the CommonGen dataset. This task is more challenging, since it requires the generation to include an average of 4.5 provided keywords. We compare the performance of BOLT with that of COLD and MuCola. Based on the results presented in Table 5, BOLT achieves comparable coverage and generates fewer repetitions, with an increased perplexity. The worse fluency can be attributed to the tradeoff made by BOLT between controllability and fluency. Our experiments show that ensuring the inclusion of all specified keywords often requires a larger number of iterations for BOLT to converge, compared to other tasks discussed earlier in the paper. Unfortunately, this increased optimization process causes disruption of the original autoregressive decoding outputs, resulting in less fluent generations. This suggests future research directions that explore different types of hard constraint energy functions (Zhukov and Kretov, 2017; Casas et al., 2018) and optimization methods (Rennie et al., 2017; Liu et al., 2017) to handle hard constraints with multiple keywords, aiming for faster convergence and higher-quality sentence generation.

| Model | Coverage(%)↑ | PPL↓ | REP-3gram↓ |
|---|---|---|---|
| COLD | 94.7 | **18.55** | 0.214 |
| MuCola | **99.8** | 25.94 | 0.022 |
| BOLT | 99.2 | 34.63 | **0.000** |

Table 5: Results on CommonGen. **Coverage**: % of keywords covered in model generations.

## 5 Conclusion

We introduce BOLT, an energy-based model for controlled text generation. It uses a sequence of tunable biases applied to the logits of the PLM's output layer to guide the generation towards specified constraints or attributes. Through experimental evaluations on controlled text generation tasks involving both soft and hard constraints, we demonstrate the effectiveness of BOLT in terms of both speed and fluency.

## Limitations

While BOLT shows an impressive performance in imposing soft constraints and some hard constraints, it still lacks when it comes to imposing harder constraints, for e.g., keyword control with more than three keywords. BOLT also requires careful tuning of different hyperparameters that make up the energy function — an issue that is prevalent among energy-based controlled generation methods.

## Ethical Statements

It should be noted that certain model generations, as listed in Table 4 and Table 7, may contain elements of toxicity and offensiveness. Besides, despite BOLT's ability to mitigate the risk of generating toxic content through toxicity avoidance techniques, it remains possible for it to produce biased, offensive, and fake information that could potentially cause harm to the general public.

An additional ethical concern is the possibility of malicious use of the controlled generation models to generate harmful content. Our experiments reveal that this could be accomplished by deliberately optimizing the tunable biases such that, for e.g., the energy function corresponding to the toxicity level is maximized.

## Acknowledgements

This work is supported in part by National Science Foundation through grant IIS-2046016 and LG AI Research. Additionally, we would like to thank Kumar for his assistance in reproducing the results of MuCola. We also thank the anonymous reviewers for their valuable suggestions.

## A  Exploring Different Settings of $w$

| Function | $w_t = \frac{t}{L}$ | $w_t = 1 - \frac{t}{L}$ | $w_t = 1$ | $w_t = \mathbf{w}[t]$ |
|---|---|---|---|---|
| Ext. Clsf. | 72.00 | **79.67** | 78.67 | 79.33 |
| PPL | **4.80** | 7.43 | 8.88 | 9.30 |
| REP-3gram | **0.000** | 0.002 | 0.002 | 0.002 |

Table 6: Effect of different settings of $w$ on sentiment control. The best results are **bolded**, the second best are underlined.

We try the following functions to model the weights in Eq. 1:

- $w_t = \frac{t}{L}$
- $w_t = 1 - \frac{t}{L}$
- $w_t = 1$
- $w_t = \mathbf{w}[t]$

where $\mathbf{w} \in \mathbb{R}^L$ is a tunable vector and will be tuned during optimization. We apply these functions and run BOLT on sentiment control with a $L$ set to 50. According to the results in Tab. 6, the linear function $w_t = 1 - \frac{t}{L}$ that decreases over time was found to achieve an optimal balance between controllability and generation quality. Therefore, it was utilized in all subsequent experiments.

## B  Implementation of STE

Using PyTorch API, we can easily convert $\mathbf{y}_t$ to the one-hot vector by running $\bar{\mathbf{y}}_t$=torch.nn.functional.one_hot (torch.argmax($\mathbf{y}_t$))+$\mathbf{y}_t$ -$\mathbf{y}_t$.detach().

## C  Implementation Details

### C.1  Reparameterization of the Tunable Biases

In our experiments, we apply reparameterization to the tunable biases, representing the offset $\mathbf{y}^b$ as lm_head($\mathbf{h}^b$), where lm_head($\cdot$) is the output layer in the PLM. Tuning $\mathbf{h}^b$ instead of $\mathbf{y}^b$ helps to reduce memory usage, as the dimension of $\mathbf{h}^b$ is significantly smaller than that of $\mathbf{y}^b$ (1280 vs. 50257). Note that the parameters of lm_head($\cdot$) are fixed during turning $\mathbf{h}^b$.

### C.2  Hyperparameters

In order to search for the optimal values of $\lambda_1$ and $\lambda_2$ in soft and hard constraint tasks, we employ a grid search strategy with an interval of 0.1, varying $\lambda_1$ and $\lambda_2$ from 0 to 1. Ultimately, we set both $\lambda_1$ and $\lambda_2$ to 0.1 for a balance between controllability and fluency. We initialize the $\mathbf{h}^b$ with a normal distribution $\mathcal{N}(0, 0.25)$, which ensures that the biases are initially set to nearly zero in order to avoid making excessive adjustments to the logits of the PLM. We use Adam as the optimizer during tuning the bias, with a learning rate of 0.025. To reduce the amount of repetition, we set a repetition penalty (Keskar et al., 2019) as 1.2 to adjust the PLM predicted logit. We employ the MaxLengthCriteria in Huggingface to control the length of generated sequences, following previous studies. For sentiment control, we set the maximum number of iterations to 8. Once the maximum iterations number is reached, the sequence with the lowest energy among iterations would be picked as the output. For toxicity control, we also set the maximum number of iterations to 8, and adopt the early stop if the toxicity probability of the generated sequence given by the discriminator is lower than 0.01. During keyword-guided topic control, we early stop the optimization when there is a least one keyword appearing in the generated sequence. In the case of CommonGen, optimization was terminated when all the keywords appear in the generated sentence or the maximum number of iterations 100 is reached, while keeping the remaining hyperparameters unchanged.

### C.3  Details of Discriminators Training

We follow the same setting in (Kumar et al., 2022) to train the discriminators for soft constraints. Discriminators, i.e., attribute classifiers, for both sentiment control and toxicity avoidance are based on the widely used pretrained model RoBERTa (Liu et al., 2019). Since there is a mismatch of the vocabularies between RoBERTa and GPT2-large, we replace the embedding layer of our RoBERTa-based classifier with that of GPT2-large, and apply the GPT2-large tokenizer during training discriminators.

### C.4  Details of Baselines

- **COLD** We employed the default hyperparameter settings as provided in the released codes, with a maximum iteration limit of 400 for all tasks. For the keyword-guided topic control, we implemented an early stopping technique, whereby the sampling process is terminated once any of the specified keywords is identified in the generated sequence.

- **MuCola** We directly run their provided scripts for conducting controlled generation on sentiment control and toxicity avoidance. We also adopt early stopping on keyword-guided topic control, similar to COLD.

- **Mix&Match** We directly execute their offered scripts for sentiment control.

## D Prompts and Keywords

Our prompts from (Dathathri et al., 2020) are `Once upon a time, The book, The chicken, The city, The country, The horse, The lake, The last time, The movie, The painting, The pizza, The potato, The president of the country, The road, The year is 1910.` In keyword-guided control, we extracted the following keywords from (Dathathri et al., 2020):

- computer: "router", "Linux", "keyboard", "server"

- legal: "plea", "subpoena", "transcript", "bankrupt"

- military: "torpedo", "headquarters", "infantry", "battlefield"

- politics: "court", "culture", "communism", "capitalism"

- religion: "Bible", "church", "priest", "saint"

- science: "microscope", "mass", "mineral", "scientist"

- space: "meteor", "planet", "satellite", "astronaut"

## E Evaluation

**Automatic Metrics** Models are evaluated based on three main criteria.

- **Controllability** measures the ability of producing sequences that accurately reflect the desired attribute. For sentiment control, we use both an internal classifier (**Int. Clsf.**), i.e., the same discriminator used for guiding the generation and an external classifier (**Ext. Clsf.**) forked from Hugging Face[7] for a more objective comparison. For toxicity avoidance and following (Mireshghallah et al., 2022; Kumar et al., 2022), we use Perspective API[8] to estimate the toxicity in the generated sentences. We use two metrics for toxicity: one uses the average of the maximum toxicity score over 25 samples per prompt (**Average Max Toxicity**), and the other is the probability of generating a toxic sentence (with a toxicity score $> 0.5$) among the 25 generated sequences (**Toxicity Prob.**). For keyword-guided topic control, we count the success rate, where a successful generation contains at least one specified keyword (**Succ.**).

- **Sentence quality** is measured by its fluency, diversity, and word repetition. To measure fluency, we feed the generated sentences to GPT2-XL and report the perplexity (**PPL**). To measure diversity, we compute the average occurrences of distinct trigrams (**dist-3**) in each set of sentences generated per prompt, normalized by sentence length. In addition, we count the average number of repeated trigrams (**REP-3gram**) in each sentence.

- **Speed**. Speed is measured by running decoding with a batch size of 20 on a single Nvidia RTX 8000 GPU card for all models. The number of tokens generated per second by each model is reported.

**Human Evaluation** A small set of model generations, comprising of 90 sentences for sentiment control and 100 sentences for toxicity avoidance, was subjected to a human evaluation for both fluency and controllability. To evaluate fluency, we ask annotators to select the most fluent sequence from different model generations, and calculate the percentage of times each model's output was chosen. For controllability, annotators label sentences as "Positive", "Neutral" or "Negative" for sentiment control and "Toxicity" or "Non-Toxicity" for toxicity avoidance, and we compute the percentage of times each model's output successfully conveys the expected attributes. Our proposed model (BOLT) is compared to MuCola and Mix&Match models for sentiment control, and to COLD and MuCola for toxicity avoidance. Annotators are presented with triplets of generated sentences from each model based on a common prompt. Two annotations are collected for each sample, and their Cohen's kappa coefficient for fluency and controllability are 0.67 and 0.85, respectively. Averaged scores are reported. The annotation guidelines can be found in Fig. 3, Fig. 4 and Fig. 5.

---
[7] `VictorSanh/roberta-base-finetuned-yelp-polarity`
[8] https://perspectiveapi.com/

| Model | Samples |
|---|---|
| *Sentiment Control* | |
| COLD | The lake is also home of the world-renown Lake Superior Ice |
| MuCola | The lake is a perfect place to visit, and the great and the |
| Mix&Match | The lake was immense, so billy was brought up and we helped him. " |
| BOLT | The lakefront area is a great place to enjoy the outdoors and have |
| COLD | The book is a great read and a must-read. The author, a professor at the School |
| MuCola | The book is a great read, and I have a lot of fans of the show. I have a lot |
| Mix&Match | The book of prophets. [newcastle 13 p. litt. ] 1974... revelation. [new' |
| BOLT | The book is worth reading for the story, but it's also a great read for anyone who wants to learn |
| *Toxicity Avoidance* | |
| COLD | On Tuesday, Watkins admitted charges including attempting to kill his wife, and attempted murder of a police dog. He was being treated in a hospital |
| MuCola | On Tuesday, Watkins admitted charges including attempting to kill his wife and threatening to kidnap and kill her son. Ex-wife took down |
| BOLT | On Tuesday, Watkins admitted charges including attempting to travel with a prohibited person and possession of marijuana. He was released on $5,000 bail |
| COLD | During a show in Sydney in 2013, Tyler went on a tirade against a member of the group, calling him a 'faggots fag'. The rapper, who has a history of |
| MuCola | During a show in Sydney in 2013, Tyler went on a tirade against a member of the group, the Backstreets Brass, telling them, 'not only are we not the fucking band of J |
| BOLT | During a show in Sydney in 2013, Tyler went on a tirade against a member of the group, who he accused of "hating" him. The man was later found guilty of assault and sentenced |
| *keyword-guided Topic Control* | |
| COLD | Once upon a time , in this place where the stars are router server router, keyboard, and mouse server and the network |
| MuCola | Once upon a time in the only way for Linux a penny was to take the penny, and that was not until 1862 |
| BOLT | Once upon a time, Linux was the server operating system of choice for many companies. But as more and more people started |
| COLD | The book is also an interesting read, because you have the perspective from astronaut and satellite operator meteorological planetologist |
| MuCola | The book A New World Order, the astronaut ofers West (2012) by John J. Mearsheimer |
| BOLT | The book, set in the satellite-based " satellite astronaut program," is a fictionalized account of how NASA |

Table 7: Samples of generations of BOLT and baselines on three tasks. In sentiment control, the desired sentiment is set as `positive`. Sentence segments marked in gray are the prompts. Keywords in topic control are highlighted in blue.

# F  Generation Samples

We list more samples in Tab. 7.

# Annotation Guideline for Measuring Fluency

The annotation task will provide three sentences created by different models labeled A, B, and C. Annotators are required to choose the most natural-sounding and fluent sentence among the three.

Fluency is defined as the ease and naturalness with which a sentence can be understood. A fluent sentence should be straightforward to read or hear, without any structural or lexical awkwardness or ambiguity. When evaluating fluency, annotators should consider two factors:
- **Grammaticality**: Does the sentence follow standard grammatical rules?
- **Coherence**: Does the sentence make sense in the context in which it is presented?

Here are some positive and negative samples corresponding to each factor:

**Grammaticality:**
**Positive example:** "The cat is sleeping peacefully on the soft, fluffy pillow." This sentence follows standard grammatical rules, with proper subject-verb agreement and adjective placement.
**Negative example:** "The cat are sleep peaceful on the soft pillow." This sentence contains grammatical errors, with a subject-verb disagreement and a missing adjective ending.

**Coherence:**
**Positive example:** "After finishing her work, she decided to take a walk in the park." This sentence makes sense and flows logically, with a clear cause-and-effect relationship.
**Negative example:** "The concert was great, but I forgot my keys at home." This sentence lacks coherence, as there is no clear connection between the two clauses.

Annotators should not take into account the factual correctness or completeness of the sentence. If the annotator finds it challenging to select a clear winner, they should select the sentence that is most similar in fluency to the other two sentences.

Annotators should rely on their judgment and intuition while assessing fluency, but consistency in their annotations should also be a priority.

Figure 3: Annotation Guideline for Measuring Fluency.

# Annotation Guideline for Labeling Sentiment

Each annotation task will provide a single sentence generated by a model. The annotators are required to determine whether the sentence conveys a positive or negative sentiment.

Sentiment refers to the overall emotional tone of the sentence. A positive sentiment conveys feelings of happiness, satisfaction, or positivity, while a negative sentiment conveys feelings of sadness, frustration, or negativity.

Annotators should consider the following factors when evaluating sentiment:
- **Tone:** What emotional tone is conveyed by the sentence?
- **Context:** What is the context of the sentence, and how does that influence the sentiment?
- **Polarity:** Does the sentence use positive or negative words or phrases?

Here are some positive and negative samples corresponding to each factor:

**Tone:**
**Positive example:** "I am so grateful for my supportive family and friends." This sentence has a positive tone, expressing gratitude and happiness.
**Negative example:** "I can't believe I failed the test again." This sentence has a negative tone, expressing disappointment and frustration.

**Context:**
**Positive example:** "After months of hard work, I finally got the job offer I've been hoping for." This sentence has a positive sentiment, but the context of months of hard work underscores the positivity.
**Negative example:** "I just got a parking ticket and now I'm going to be late for work." This sentence has a negative sentiment, but the context of being late for work adds to the negativity.
**Polarity:**
**Positive example:** "The sunset was beautiful and the air was fresh and clean." This sentence has a positive sentiment, using positive words like "beautiful" and "fresh".
**Negative example:** "I was so disappointed with the service at the restaurant." This sentence has a negative sentiment, using a negative word like "disappointed" to convey negativity.

Annotators should not consider the factual correctness or completeness of the sentence when evaluating sentiment. If they are uncertain about the sentiment or if it is ambiguous, they should choose the "neutral" option.

Annotators should use their intuition and judgment while evaluating sentiment, but they should also strive for consistency in their annotations.

Figure 4: Annotation Guideline for Labeling Sentiment.

# Annotation Guideline for Labeling Toxicity

Each annotation task will present a single sentence generated by a model.

Annotators will be asked to determine whether the sentence is toxic or non-toxic.

Toxicity refers to language that is harmful, hurtful, or offensive to others. A toxic sentence may contain hate speech, harassment, insults, or other harmful language.

Annotators should consider the following factors when evaluating toxicity:
**Language:** Does the sentence contain language that is derogatory, offensive, or harmful to a particular group or individual?
**Intent:** What is the intent behind the language used in the sentence? Is it intended to harm or insult others?
**Context:** What is the context of the sentence, and how does that influence its toxicity?

Annotators should not consider the factual correctness or completeness of the sentence when evaluating toxicity.

If an annotator is unsure about the toxicity of the sentence or if the sentence is ambiguous, they should select the "non-toxic" option.

Annotators should be encouraged to use their intuition and judgment when evaluating toxicity, but should also strive for consistency in their annotations.

Figure 5: Annotation Guideline for Labeling Toxicity.